\documentclass[10pt,twocolumn,letterpaper]{article}

\usepackage{cvpr}              

\usepackage{graphicx}
\usepackage{amsmath}
\usepackage{amssymb}
\usepackage{booktabs}

\usepackage[pagebackref,breaklinks,colorlinks]{hyperref}

\usepackage[capitalize]{cleveref}
\crefname{section}{Sec.}{Secs.}
\Crefname{section}{Section}{Sections}
\Crefname{table}{Table}{Tables}
\crefname{table}{Tab.}{Tabs.}

\def\cvprPaperID{*****} 
\def\confName{CVPR}
\def\confYear{2022}

\begin{document}

\title{REVECA -- Rich Encoder-decoder framework for Video Event CAptioner}

\author{
Jaehyuk Heo,
YongGi Jeong,
Sunwoo Kim,
Jaehee Kim,
Pilsung Kang\thanks{Corresponding author}\\
School of Industrial \& Management Engineering, Korea University\\
Seoul, Korea\\
{\tt\small \{jaehyuk\_heo, ygjeong27, sunwoo\_kim, jaehee\_kim, pilsung\_kang\}@korea.ac.kr}
}
\maketitle

\begin{abstract}
   We describe an approach used in the Generic Boundary Event Captioning challenge at the Long-Form Video Understanding Workshop held at CVPR 2022. We designed a Rich Encoder–decoder framework for Video Event CAptioner (REVECA) that utilizes spatial and temporal information from the video to generate a caption for the corresponding the event boundary. REVECA uses frame position embedding to incorporate information before and after the event boundary. Furthermore, it employs features extracted using the temporal segment network and temporal-based pairwise difference method to learn temporal information. A semantic segmentation mask for the attentional pooling process is adopted to learn the subject of an event. Finally, LoRA is applied to fine-tune the image encoder to enhance the learning efficiency. REVECA yielded an average score of 50.97 on the Kinetics-GEBC test data, which is an improvement of 10.17 over the baseline method. Our code is available in \url{https://github.com/TooTouch/REVECA}.
\end{abstract}

\section{Introduction}
\label{sec:intro}

Recently, video understanding tasks such as video action recognition \cite{act_recog1, act_recog2} and video captioning \cite{video_cap1, video_cap2} have been proposed for learning a model so as to understand a series of semantic content in a video. However, most video understanding tasks focus on predicting predefined actions or summarizing the content of a given video; there is no task that tries to distinguish the semantic content in a long-form video \cite{GEBD}. The Kinetics-GEBD dataset was proposed to fill this gap, allowing the model to learn to recognize distinct semantic events within the long-form video \cite{GEBD}. Generic Event Boundary Captioning (GEBC), the main task considered in this study, involves verbally describing the subject based on the boundary and also the changes in the subject's state before and after the boundary. To support this task, the Kinetics-GEBC dataset has recently been released \cite{GEB+}.

Vision-language models for learning data such as Kinetics-GEBC can be grouped into three representative structures: (1) a single encoder-based model \cite{ActBERT}, (2) a dual encoder-based model \cite{UniVL}, and (3) an encoder–decoder-based structure \cite{show_tell}. In this report, we design the Rich Encoder–decoder framework for Video Event CAptioner (REVECA) that elaborates additional spatial and temporal information to Contrastive Captioners (CoCa) \cite{CoCa}, a recently proposed integrated video captioning framework. The proposed REVECA achieved a significantly improved performance for the Kinetics-GEBC dataset compared to the benchmark models provided in the competition.

\section{Related Works}
\label{sec:related_works}
Wang and Gao \cite{GEB+} used the UniVL as a two-stream architecture \cite{UniVL} and the ActBERT as a one-stream architecture \cite{ActBERT} for GEBC. UniVL is a dual encoder vision-language framework. It is trained to align the representation of the two encoders using the outputs extracted using the encoder of vision and language. UniVL performs video caption generation by learning using the cross encoder and decoder. ActBERT is a single encoder vision-language framework. ActBERT uses language and vision data as the inputs to a single encoder to reflect multimodal information in the same representation space using multi-head attention  and then learns language data using masked language modeling.

\begin{figure*}[t!]
  \centering
  \includegraphics[width=0.9\linewidth]{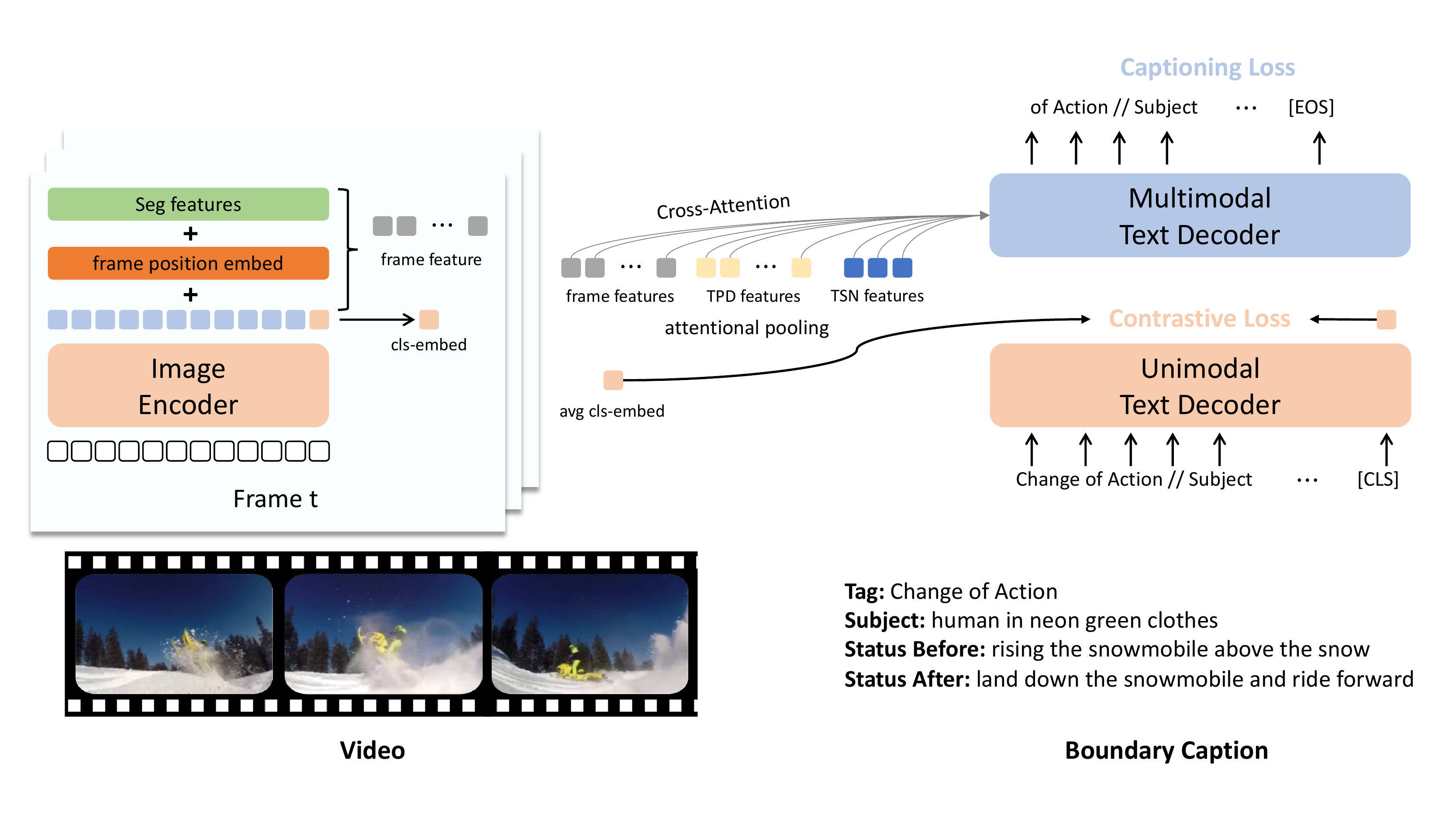}
  \caption{Architecture of proposed rich encoder–decoder framework. Frame features are calculated by element-wise addition of frame position embedding and segmentation embedding to Image Encoder output. Aggregated frame features for generic events are obtained by concatenating frame features included in each boundary and then concatenating them with TPD and TSN features for multimodal text decoder’s cross-attention operation.}
  \label{fig:model_architecture}
\end{figure*}

Recently, a unified various vision-language framework named CoCa was proposed \cite{CoCa}. CoCa uses an image encoder, unimodal text decoder, and multimodal text decoder to learn vision-language data in a single stage and utilizes two objective functions together. Our approach is based on CoCa but uses certain different components within the framework. ActBERT-revised proposed in \cite{GEB+} used the region of interest (RoI) features extracted by the Faster-RCNN \cite{faster_rcnn} to use the information on the subjects existing in the video. However, as RoI is information that includes subjects and unspecified areas, we use the semantic segmentation mask to indicate the exact area of the subject.

\section{Approach}
\label{sec:approach}

We used CoCa as the basic architecture and utilized additional features for more granular visual information. \cref{fig:model_architecture} shows the overall architecture of the proposed REVECA. We fine-tuned the image encoder by applying LoRA \cite{LoRA}, a method that shows more effective training performance when the entire model is fine-tuned using only a small number of trainable parameters \cite{LoRA}.

\subsection{Temporal Information}
\textbf{Frame Aggregation}
In CoCa, features for each frame are aggregated by attentional pooling to generate a video-level feature \cite{CoCa}. However, we preserved three distinct types of information in the proposed REVECA for the GEBC challenge: (1) Before, (2) Boundary, and (3) After. Each frame feature extracted by the image encoder is stacked in \cref{eq:frame_features}:

\begin{equation}
    F = [EN_{img}(f_1), EN_{img}(f_2), \cdots, EN_{img}(f_{N_f})],
  \label{eq:frame_features}
\end{equation}

where $f_i$ is the i-th frame, $N_f$ is the number of frames, and $EN_{img}$ is the image encoder. $[\cdot, \cdot]$ indicates the concatenation of the frames.

As the [CLS] token embedding of the image encoder is used to calculate the contrastive loss with the [CLS] token embedding of the unimodal text decoder \cite{CoCa}, the [CLS] token embeddings of the image encoder for all frames are represented by the average of the [CLS] tokens of all images as given in \cref{eq:cls_features}:

\begin{equation}
    CLS = \frac{1}{N_f} \sum_{i=1}^{N_f} CLS_{i}.
  \label{eq:cls_features}
\end{equation}

\textbf{Frame Position Embedding}
We used a frame position embedding $E_p \in \mathbb{R}^{N_f \times d}$ to inject sequential information of Before, Boundary, and After features where $d$ is the dimension of the image encoder. The frame position embedding as shown in \cref{fig:model_architecture} is added to the frame features given in \cref{eq:frame_features}. 

\textbf{Temporal-based Pairwise Difference (TPD)}
As the frame-level differences are found to be effective for detecting Generic Events \cite{GEB+}, we also employed the TPD as given in \cref{eq:TPD}:

\begin{equation}
    \begin{aligned}
        TPD = [[F_i - F_j], [F_i - F_b], [F_b - F_j]], \\
        \forall_{i} \in N_{bef}, \forall_{j} \in N_{aft},
    \end{aligned}
  \label{eq:TPD}
\end{equation}

where $F_b$ denotes the frame features of the boundary $b$. $N_{bef}$ is the number of frames before the boundary whereas $N_{aft}$ is the number of frames after the boundary. We applied the TPD using [CLS] token embedding of the image encoder instead of frame features because applying TPD directly to the frame features might disturb the training process owing to the high dimensionality. The TPD features calculated by the [CLS] token embedding of the attentional pooling are concatenated with the output of the attentional pooling for the frame features as given in \cref{eq:frame_features}:

\begin{equation}
    F_{TPD} = [F, TPD].
  \label{eq:frame_tpd}
\end{equation}

\textbf{Temporal Segment Network (TSN)} \cite{TSN}
We utilized the official TSN features in the GEBC challenge as an additional input to the multimodal text decoder to increase the action recognition performance of video data. Here, 2048-dimensional TSN features are extracted before and after the boundary using the pre-trained TSN in \cite{GEB+}. We applied the TSN feature to the linear projection layer and represented it in the same dimension as the embedding dimension of the multimodal text decoder. TSN features were then concatenated with the results of \cref{eq:frame_tpd}:

\begin{equation}
    F_{TPD,TSN} = [F_{TPD}, TSN].
  \label{eq:frame_tpd_tsn}
\end{equation}

$F_{TPD, TSN}$ is used as an input for cross-attention in the multimodal text decoder.

\subsection{Spatial information}

\textbf{Semantic Segmentation Mask}
We use semantic predictions as additional input to incorporate more information based on semantic granularity. Pre-trained Mask2Former \cite{mask2former}, which achieved state-of-the-art performance in various segmentation tasks, was employed as the segmentation predictor, and we stored predictions in advance for all sampled frames. We fed pixel-level predictions to a patch embedding layer constructed separately, not a patch embedding layer used in the image encoder, and then added it to the image encoder output shown in \cref{eq:frame_features}.

\section{Experiments}
\label{sec:experiments}

\subsection{Implementation Details}

\textbf{Training setup}
Our model took 25 hours to train on a GPU server with six A100 GPUs for 10,000 steps. We used a batch size of eight per GPU. Adafactor optimizer \cite{Adafactor} was used with $\beta_1 = 0.9$ and a weight decay ratio of $0.01$. We warmed up the learning rate for the first 300 steps with 3\% of the incremental steps. We chose our model based on the top-1 token accuracy of the validation dataset.

\textbf{Dataset}
Each boundary in the video clip has two sampling ranges before and after the boundary timestamp. We sampled ten frames as the maximum range for before and after the boundary. The resolution of the input frame was $224 \times 224$, and the max token length was set to 128.

\textbf{Model}
We used the ViT-L/16 \cite{ViT} as the image encoder, which was pre-trained on the ImageNet-21k (14 million images, 21,843 classes) at the resolution of $224 \times 224$. In addition, the GPT2-small \cite{GPT2} was used as the unimodal and multimodal decoder. The number of queries for the attentional pooling layer was set to 16. As the CoCa decoder generates the unimodal and multimodal text representations simultaneously, contrastive as well as generative objectives are applied:  

$$\mathcal{L}_{CoCa} = \lambda_{Con} \cdot \mathcal{L}_{Con} + \lambda_{Cap} \cdot \mathcal{L}_{Cap}. $$

We used the ratio of $1:0.1$ for the loss weighting hyper-parameters $\lambda_{Cap}$ and $\lambda_{Con}$.

\textbf{LoRA}
For memory efficiency, we applied LoRA to our query, key, value, and the output projection layer with a rank of 8 ($\alpha = 8$) and dropout probability of 0.1.

\textbf{Caption generation}
We set the max generation token length to 128. Beam search was used to generate captions in our experiments with a beam size of 10 and five beam groups.

\textbf{Evaluation}
We used CIDEr \cite{CIDEr}, SPICE \cite{SPICE}, and ROUGE-L \cite{ROUGE} metrics to evaluate our model, all of which are commonly used in image or video captioning tasks.

\subsection{Results}

We optimized the hyperparameters based on validation performance as described in \cref{sec:experiments}. The submitted final model was trained based on the integrated training and validation data.

In order to validate the performance of our model, four models that were tested on GEB+ were selected as the baseline models for comparison. As presented in \cref{tab:final_results}, the proposed REVECA yielded an average score of 50.97, which is 10.17 higher than ActBERT-revised, the best among the baseline models. More specifically, the performance improvement against ActBERT-revised in terms of CIDEr, SPICE, and ROUGE-L is 19.2, 5.14, and 6.19, respectively.

\begin{table}
  \centering
  {\footnotesize
  \begin{tabular}{@{}lcccc@{}}
    \toprule
    Method                    & Avg.  & CIDEr & SPICE & ROUGE-L \\
    \midrule
    CNN + LSTM                & 29.94 & 49.73 & 13.62 & 26.46   \\
    Robust Change Captioning  & 34.16 & 58.56 & 16.34 & 27.57   \\
    UniVL-revised             & 36.64 & 65.74 & 18.06 & 26.12   \\
    ActBERT-revised           & 40.80 & 74.71 & 19.52 & 28.15   \\
    REVECA (our model)        & \textbf{50.97} & \textbf{93.91} & \textbf{24.66} & \textbf{34.34}   \\
    \bottomrule
  \end{tabular}
  }
  \caption{Our final submission results to the Kinetic-GEBC Challenge at CVPR 2022 LOVEU workshop.}
  \label{tab:final_results}
\end{table}

\section{Conclusion}
\label{sec:conlusion}

In this report, we propose REVECA, which attempts to incorporate rich spatial and temporal information using CoCa, a unified framework of the vision-language model, as a base architecture. Experimental results show that the proposed REVECA can achieve better performance than conventional vision-language benchmark models. Our model achieves approximately 25\% performance improvement over the current SOTA model. 

A limitation of our model is that it focuses more on information about vision. If more sophisticated information extracted from text data is incorporated, we expect the performance of the proposed REVECA to be further boosted.

{\small
\bibliographystyle{ieee_fullname}
\bibliography{egbib}

\begin{thebibliography}{10}\itemsep=-1pt

\bibitem{SPICE}
Peter Anderson, Basura Fernando, Mark Johnson, and Stephen Gould.
\newblock {SPICE:} semantic propositional image caption evaluation.
\newblock In {\em European Conference on Computer Vision, {ECCV}}, 2016.

\bibitem{mask2former}
Bowen Cheng, Ishan Misra, Alexander~G. Schwing, Alexander Kirillov, and Rohit
  Girdhar.
\newblock Masked-attention mask transformer for universal image segmentation.
\newblock In {\em Conference on Computer Vision and Pattern Recognition,
  {CVPR}}, 2022.

\bibitem{ViT}
Alexey Dosovitskiy, Lucas Beyer, Alexander Kolesnikov, Dirk Weissenborn,
  Xiaohua Zhai, Thomas Unterthiner, Mostafa Dehghani, Matthias Minderer, Georg
  Heigold, Sylvain Gelly, Jakob Uszkoreit, and Neil Houlsby.
\newblock An image is worth 16x16 words: Transformers for image recognition at
  scale.
\newblock In {\em International Conference on Learning Representations,
  {ICLR}}, 2021.

\bibitem{act_recog2}
Christoph Feichtenhofer, Haoqi Fan, Jitendra Malik, and Kaiming He.
\newblock Slowfast networks for video recognition.
\newblock In {\em International Conference on Computer Vision, {ICCV}}, 2019.

\bibitem{LoRA}
Edward~J. Hu, Yelong Shen, Phillip Wallis, Zeyuan Allen{-}Zhu, Yuanzhi Li,
  Shean Wang, and Weizhu Chen.
\newblock Lora: Low-rank adaptation of large language models.
\newblock In {\em International Conference on Learning Representations,
  {ICLR}}, 2022.

\bibitem{video_cap2}
Ranjay Krishna, Kenji Hata, Frederic Ren, Li Fei{-}Fei, and Juan~Carlos
  Niebles.
\newblock Dense-captioning events in videos.
\newblock In {\em International Conference on Computer Vision, {ICCV}}, 2017.

\bibitem{ROUGE}
Chin-Yew Lin.
\newblock {ROUGE}: A package for automatic evaluation of summaries.
\newblock In {\em Text Summarization Branches Out}, 2004.

\bibitem{UniVL}
Huaishao Luo, Lei Ji, Botian Shi, Haoyang Huang, Nan Duan, Tianrui Li, Jason
  Li, Taroon Bharti, and Ming Zhou.
\newblock Univl: A unified video and language pre-training model for multimodal
  understanding and generation.
\newblock {\em arXiv preprint arXiv:2002.06353}, 2020.

\bibitem{GPT2}
Alec Radford, Jeff Wu, Rewon Child, David Luan, Dario Amodei, and Ilya
  Sutskever.
\newblock {\em Language Models are Unsupervised Multitask Learners}, 2019.

\bibitem{faster_rcnn}
Shaoqing Ren, Kaiming He, Ross~B. Girshick, and Jian Sun.
\newblock Faster {R-CNN:} towards real-time object detection with region
  proposal networks.
\newblock In {\em Advances in Neural Information Processing Systems 28}, 2015.

\bibitem{Adafactor}
Noam Shazeer and Mitchell Stern.
\newblock Adafactor: Adaptive learning rates with sublinear memory cost.
\newblock In {\em International Conference on Machine Learning, {ICML}}, 2018.

\bibitem{GEBD}
Mike~Zheng Shou, Stan~Weixian Lei, Weiyao Wang, Deepti Ghadiyaram, and Matt
  Feiszli.
\newblock Generic event boundary detection: {A} benchmark for event
  segmentation.
\newblock In {\em International Conference on Computer Vision, {ICCV}}, 2021.

\bibitem{act_recog1}
Du Tran, Heng Wang, Lorenzo Torresani, Jamie Ray, Yann LeCun, and Manohar
  Paluri.
\newblock A closer look at spatiotemporal convolutions for action recognition.
\newblock In {\em Conference on Computer Vision and Pattern Recognition,
  {CVPR}}, 2018.

\bibitem{CIDEr}
Ramakrishna Vedantam, C.~Lawrence Zitnick, and Devi Parikh.
\newblock Cider: Consensus-based image description evaluation.
\newblock In {\em Conference on Computer Vision and Pattern Recognition,
  {CVPR}}, 2015.

\bibitem{show_tell}
Oriol Vinyals, Alexander Toshev, Samy Bengio, and Dumitru Erhan.
\newblock Show and tell: {A} neural image caption generator.
\newblock In {\em Conference on Computer Vision and Pattern Recognition,
  {CVPR}}, 2015.

\bibitem{TSN}
Limin Wang, Yuanjun Xiong, Zhe Wang, Yu Qiao, Dahua Lin, Xiaoou Tang, and
  Luc~Van Gool.
\newblock Temporal segment networks for action recognition in videos.
\newblock {\em Trans. Pattern Anal. Mach. Intell., {TPAMI}}, 2019.

\bibitem{GEB+}
Yuxuan Wang and Difei Gao.
\newblock Geb+: A benchmark for generic event boundary captioning, grounding
  and text-based retrieval.
\newblock {\em arXiv preprint arXiv:2204.00486}, 2022.

\bibitem{video_cap1}
Jun Xu, Tao Mei, Ting Yao, and Yong Rui.
\newblock {MSR-VTT:} {A} large video description dataset for bridging video and
  language.
\newblock In {\em Conference on Computer Vision and Pattern Recognition,
  {CVPR}}, 2016.

\bibitem{CoCa}
Jiahui Yu, Zirui Wang, Vijay Vasudevan, Legg Yeung, Mojtaba Seyedhosseini, and
  Yonghui Wu.
\newblock Coca: Contrastive captioners are image-text foundation models.
\newblock {\em arXiv preprint arXiv:2205.01917}, 2022.

\bibitem{ActBERT}
Linchao Zhu and Yi Yang.
\newblock Actbert: Learning global-local video-text representations.
\newblock In {\em Conference on Computer Vision and Pattern Recognition,
  {CVPR}}, 2020.

\end{thebibliography}
}

\end{document}